\pdfoutput=1

\documentclass[11pt]{article}

\usepackage{acl}

\usepackage{times}
\usepackage{latexsym}

\usepackage[T1]{fontenc}

\usepackage[utf8]{inputenc}

\usepackage{microtype}
\usepackage{graphicx}
\usepackage{color,soul}
\usepackage{multirow}
\usepackage{amsmath}
\usepackage{tabularx}
\usepackage{bbding}
\usepackage{multirow}
\usepackage{hyperref}

\title{Plot Writing From \textit{\st{Scratch}} Pre-Trained Language Models
}

\author{Yiping Jin$^1$\thanks{\enspace Work done while at Knorex.}, Vishakha Kadam$^2$, Dittaya Wanvarie$^1$ \\
$^1$Department of Mathematics \& Computer Science, Chulalongkorn University, Thailand\\
$^2$Knorex, 02-129 WeWork Futura, Magarpatta, Hadapsar, Pune, India\\
\texttt{Dittaya.W@chula.ac.th} \\
\texttt{\{jinyiping, vishakha.kadam\}@knorex.com} \\
}

\begin{document}
\maketitle
\begin{abstract}
Pre-trained language models (PLMs) fail to generate long-form narrative text because they do not consider global structure. As a result, the generated texts are often incohesive, repetitive, or lack content. Recent work in story generation reintroduced explicit content planning in the form of prompts, keywords, or semantic frames. Trained on large parallel corpora, these models can generate more logical event sequences and thus more contentful stories. However, these intermediate representations are often not in natural language and cannot be utilized by PLMs without fine-tuning. We propose generating story plots using \textit{off-the-shelf} PLMs while maintaining the benefit of content planning to generate cohesive and contentful stories. Our proposed method, \textsc{S{\scriptsize CRATCH}P{\scriptsize LOT}}, first prompts a PLM to compose a content plan. Then, we generate the story's body and ending conditioned on the content plan. Furthermore, we take a generate-and-rank approach by using additional PLMs to rank the generated (story, ending) pairs. We benchmark our method with various baselines and achieved superior results in both human and automatic evaluation~\footnote{Code and data available at \url{https://github.com/YipingNUS/scratchplot-story-generation}.}.
\end{abstract}

\section{Introduction}
\label{sec:intro}

Automatic story generation aims to produce interesting stories to be read by readers or help writers develop new ideas. However, generating long-form stories is challenging because language models lack global planning~\citep{hua2020pair,tan2021progressive}, discourse coherence~\citep{bosselut2018discourse,ji2021discodvt}, and common sense knowledge~\citep{xu2020controllable,ji2020language}. While individual sentences appear fluent and logical, they do not fit together as a whole and the stories often have no clear content~\citep{see2019massively,goldfarb-tarrant-etal-2020-content}. In long text generation, repetitions are also more prevalent, which cause the stories to degrade drastically~\citep{yao2019plan}.

\begin{figure}[!t]
\centering \includegraphics[width=218pt]{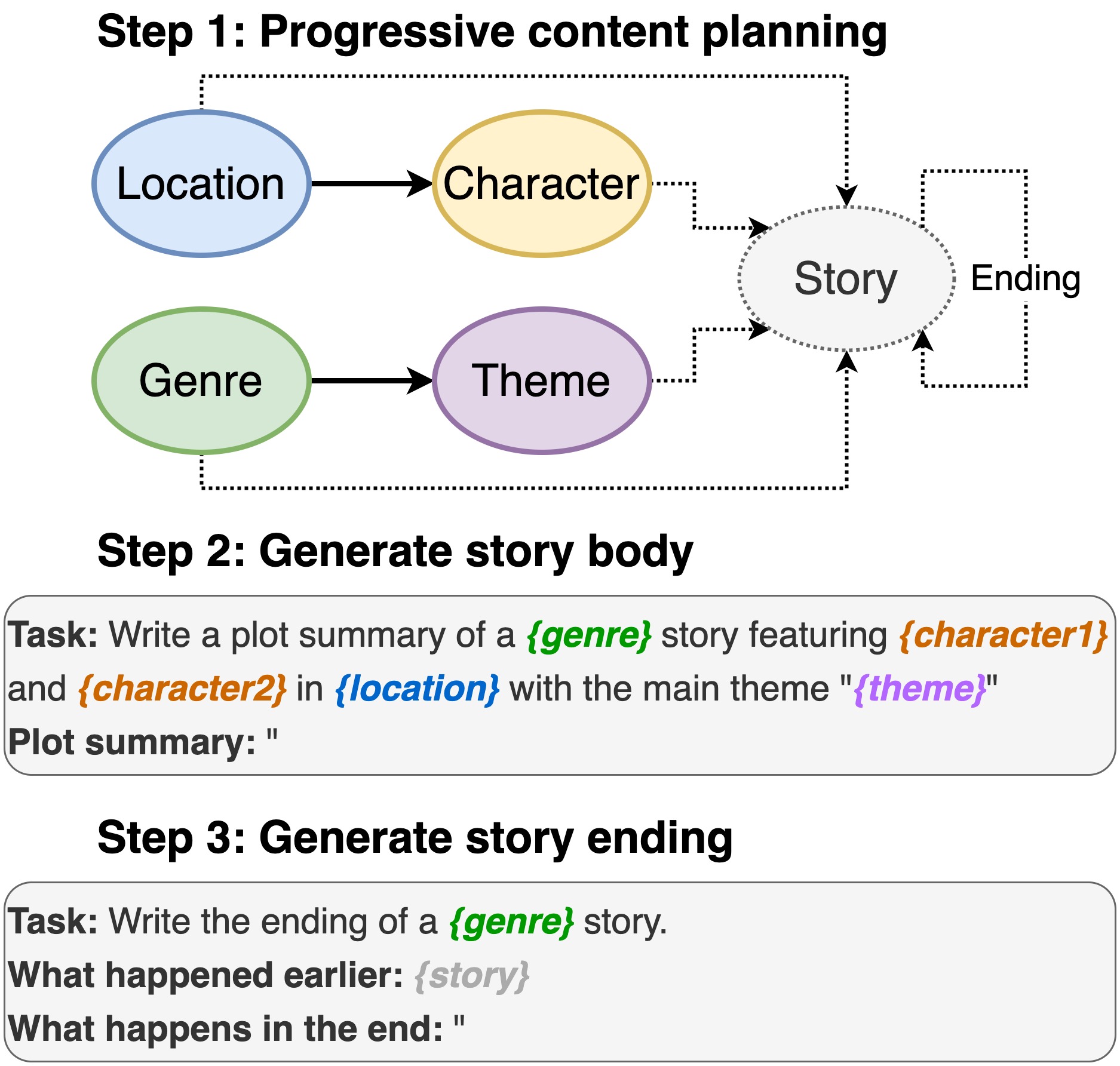}
  \caption{Overview of \textsc{S{\scriptsize CRATCH}P{\scriptsize LOT}}. We factorize the elements of a story into four attributes \{location, characters, genre, and theme\}. We first prompt a PLM to compose them sequentially, then generate the story conditioned on these attributes. When writing the ending of the story, the model additionally conditions on the previously generated story.}  
  \label{fig:overview}
\end{figure}

Interestingly, recent work in long-form story generation relied on \textit{explicit} content planning~\citep{reiter1997building}, contrary to the prevalent trend of end-to-end learning across NLP tasks. The content plan usually takes the form of prompts~\citep{fan-etal-2018-hierarchical}, keywords/keyphrases~\citep{xu-etal-2018-skeleton,yao2019plan}, semantic frames~\citep{fan-etal-2019-strategies}, or summaries~\citep{sun2020summarize}. 

These content plans are usually not in the form of natural language~\footnote{Except for using summaries as the content plan.} and cannot be understood by pre-trained language models (PLMs) without fine-tuning using parallel data. 
Another subtle problem of modeling story generation as a supervised learning task is that the model learns common sense and frequently occurring action sequences, like morning routines~\citep{fan-etal-2019-strategies}. Such an action plan may not be interesting and surprising, which are crucial characteristics of stories.

We propose generating stories using \textit{off-the-shelf PLMs without fine-tuning}. We tap on \textsc{D{\small INO}}~\citep{schick-schutze-2021-generating}, a  framework to generate datasets using instructions, to compose stories progressively. Our method, \textsc{S{\scriptsize CRATCH}P{\scriptsize LOT}}, is depicted in Figure~\ref{fig:overview}. We firstly prompt a PLM to perform content planning, including the location, characters, genre, and theme. We then generate a story conditioned on these attributes. Finally, we generate story endings and rank them. 

The proposed approach yields better stories based on various automatic and human evaluations compared with baselines fine-tuned using parallel data and a PLM with standard prompting. We only experimented with generating English stories. In principle, the approach can be applied to other languages given a reasonable PLM and task descriptions in the target language.

\section{Plot Generation From Scratch}
\label{sec:method}

\begin{figure}[!ht]
\centering \includegraphics[width=215pt]{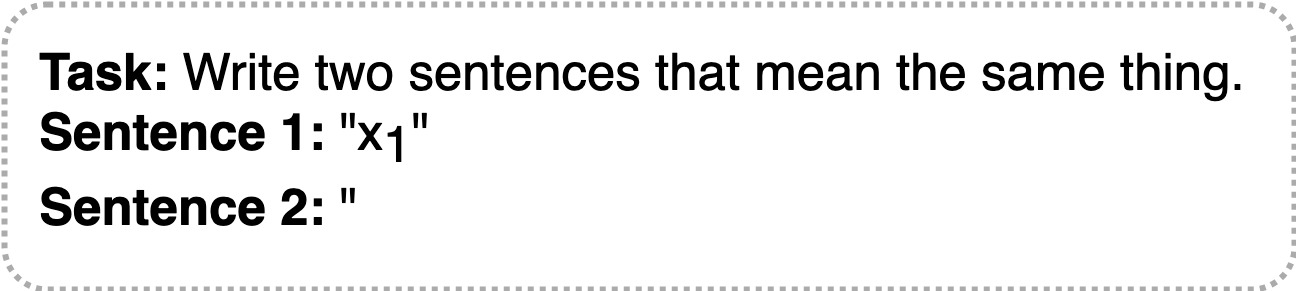}
  \caption{Task description to generate a similar sentence by incorporating the first generated sentence, $x_1$.}  
  \label{fig:dino}
\end{figure}

\textsc{D{\small INO}}~\citep{schick-schutze-2021-generating} is a framework to generate labeled NLI datasets~\citep{bowman-etal-2015-large} using a pre-trained GPT-2 model~\citep{radford2019language}. \citet{schick-schutze-2021-generating} formulated different task descriptions to generate sentence pairs for each category. Instead of generating the sentence pairs at once, they sample the first sentence $x_1$, then incorporate it into the task description to sample the second sentence, as shown in Figure~\ref{fig:dino}.

\begin{figure}[!ht]
\centering \includegraphics[width=205pt]{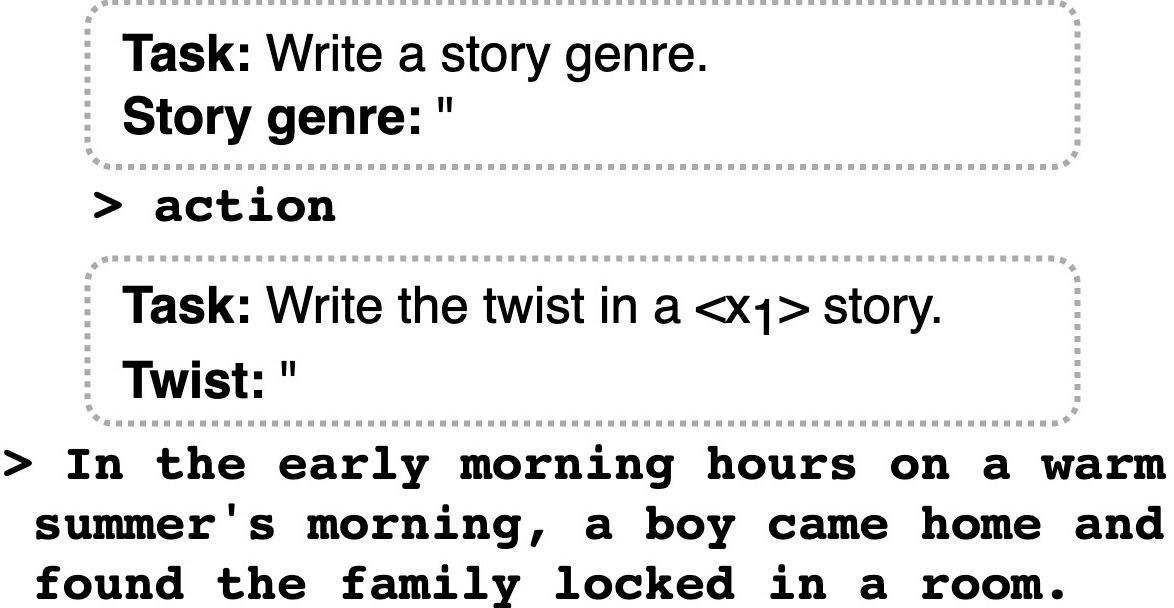}
  \caption{Task description for generating genre and theme. \textsc{<x{\scriptsize 1}>} denotes the generated genre. The example continuations are generated by GPT2-XL.}  
  \label{fig:task-descriptions}
\end{figure}

\textbf{Factorizing Elements of a Story}\quad We factorize story generation into multiple stages analogous to generating NLI sentence pairs. We define four main plot elements: location, characters, genre, and theme. These elements are not entirely independent. For example, the genre will influence the theme. We denote these dependencies using solid arrows in Figure~\ref{fig:overview}. We then use different task descriptions to sample these elements sequentially. Figure~\ref{fig:task-descriptions} shows example task descriptions to generate the genre and theme. We use paraphrases of the task descriptions, and the complete list is presented in Appendix~\ref{sec:appendix-td}. 

We apply self-debiasing~\citep{schick2021self} to generate distinct geographical units and male/female names using each task description. Meanwhile, we generate other plot elements without self-debiasing because their task descriptions complement each other. Self-debiasing rewards generated continuations which receive \textit{high} probability conditioned on one task description and \textit{low} probability conditioned on other task descriptions. For example, when generating a male name, we want it to be \textit{unlike} a female name. Specifically, we calculate the token's probability $p_y$ assigned by the PLM using each task description. The token's final logit for each task description $y$ is as follows:

\begin{equation}
\delta_y = p_y - \max_{y' \neq y}p_{y'}
\label{eq:self-debiasing}
\end{equation}

We sample one value for each plot element except for characters, where we generate a male and a female character. After sampling all plot elements, we fuse them into a single task description to generate the story, as depicted in Step 2 of Figure~\ref{fig:overview}.

\textbf{Generating Coherent Story Ending}\quad Coherent and thoughtful endings are crucial to stories. However, it is not obvious how to write the story ending with PLMs. GPT-2 does not have an \textsc{<eos>} (end of sequence) token. Therefore, \citet{schick-schutze-2021-generating} always end the task description with an opening quotation mark (as shown in Figure~\ref{fig:dino}) and generate till the first closing quotation mark. However, the PLM usually generates the first closing quotation mark after a couple of sentences, making it unsuitable for generating long-form stories. Therefore, we generate the story with a fixed length and truncate it till the last complete sentence. 

We design a separate task description to write the story ending \textit{explicitly} by providing the story body and asking the PLM to write what happens in the end (Step 3 in Figure~\ref{fig:overview}). As the story ending is usually short, we terminate at the first generated closing quotation mark following \citet{schick-schutze-2021-generating}.

We observe that PLMs sometimes ignore the task descriptions and write generic or irrelevant story endings. Therefore, we propose two methods to rank the story endings. Firstly, we use the next sentence prediction (NSP) task of BERT~\cite{devlin-etal-2019-bert} to measure the coherence between the story and the ending. Specifically, we calculate $\mathcal{P}_{NSP}(b,e)$, where $b$ denotes the story body and $e$ denotes the story ending.

%

Inspired by previous works in fact-checking with PLMs~\citep{lee-etal-2020-language,lee-etal-2021-towards}, we use the perplexity score as another metric to measure the story ending's quality. Specifically, we concatenate the story body and ending to form the input to the PLM: $X = \{x_{b_0}, ..., x_{b_B}, x_{e_0}, ..., x_{e_E}\}$, where $B$ and $E$ denote the number of tokens in the story body and ending separately. We then calculate the conditioned perplexity by

\begin{equation*}
PPL(X) = \sqrt[E]{\prod_{i=1}^{E}\frac{1}{p(x_{e_i}|x_{b_0}, ..., x_{b_B}, ..., x_{e_{i-1}})}}
\end{equation*}

Note that we use the story body tokens to condition the perplexity, but they do not contribute to the $PPL(X)$.


During inference, we sample multiple (story body, story ending) pairs and use NSP and PPL to rank them~\footnote{NSP the higher, the better. $PPL(X)$ the lower, the better.}.

\section{Experiments}
\label{sec:exp}

\textbf{Experimental Details}\quad We use the official implementation of \textsc{D{\small INO}}~\citep{schick-schutze-2021-generating}~\footnote{\url{https://github.com/timoschick/dino}} with the default GPT2-XL language model. We follow the default parameters except setting $k$=30 for top-$k$ sampling and blocking repeating trigrams during generation. For story ending ranking, we use HuggingFace~\citep{wolf-etal-2020-transformers} \texttt{bert-base-uncased} checkpoint to calculate the NSP probability and \texttt{gpt2} (base) to calculate the perplexity. 

We perform simple post-processing to clean or filter the continuations, such as removing tailing punctuations and filtering continuations that repeat words from the prompt or contain 1st or 2nd person pronouns. The story body must also contain some plot elements to ensure it is contentful and respects the task description. The post-processing is detailed in Appendix~\ref{sec:appendix-postprocessing}.

We generate plot elements \textit{offline} in batches and store them. When generating stories, we randomly sample each type of plot element and combine them to form a content plan~\footnote{We sample plot elements following the same sequence in Figure~\ref{fig:overview} to preserve the dependency among them.}. Table~\ref{tab:params} presents the detailed parameters used for each type of generation.

\begin{table}[!htbp]
\centering
  \begin{tabular}{lccc}
    \hline
    \textbf{Element} & \textbf{num} & \textbf{min\_len} & \textbf{max\_len}\\
    \hline
	Location & 20 & 1 & 5 \\
	Cast & 10 & 1 & 5 \\
	Genre & 20 & 1 & 5 \\
	Theme & 10 & 5 & 25 \\\hline
	Story Body & 30 & - & 100 \\
	Story Ending & 10 & 10 & 50 \\
	\hline
  \end{tabular}
  
\caption{Hyperparameters for generation. For (location, genre, story body), \textbf{num} indicates the number of continuations to generate per \textit{task description}. For (cast, theme, story ending), \textbf{num} is the number of continuations per \textit{input} \textsc{<x{\scriptsize 1}>} and \textit{task description} combination. Please refer to Appendix~\ref{sec:appendix-td} for the number of task descriptions for each plot element.}
\label{tab:params}
\end{table}

\textbf{Baselines}\quad We compare with the following three conditional story generation baselines. 

\begin{itemize}
  \item \textbf{Fusion}~\citep{fan-etal-2018-hierarchical}~\footnote{\url{https://github.com/pytorch/fairseq/tree/main/examples/stories}}: A seq2seq model with a convolutional encoder and a self-attention decoder generating stories conditioned on a prompt. We use the official checkpoint, which is fine-tuned on the {W{\scriptsize RITING}P{\scriptsize ROMPTS}} dataset with 300k prompt-story pairs.
  \item \textbf{Plan-and-write}~\citep{yao2019plan}~\footnote{\url{https://bitbucket.org/VioletPeng/language-model}}: A bidirectional gated recurrent unit (BiGRU) seq2seq model that first predicts the storyline (as specified by a sequence of keywords) from the title. It then generates the story conditioned on both the title and the storyline. We train the model on ROCStories dataset~\citep{mostafazadeh-etal-2016-corpus} for 280 epochs till convergence and follow the default hyper-parameters in the official repository.
  \item \textbf{ProGen}~\citep{tan2021progressive}~\footnote{\url{https://github.com/tanyuqian/progressive-generation}}: A multi-stage BART seq2seq model~\citep{lewis-etal-2020-bart} using salient keywords as intermediate representations. We use a two-stage seq2seq architecture, where the first seq2seq model takes the input keywords and generates a refined intermediate representation containing keywords with finer-grained details. The second seq2seq model then uses it as input and generates the final story. We fine-tune a BART-base model for both stages using 1k examples randomly sampled from the {W{\scriptsize RITING}P{\scriptsize ROMPTS}} dataset following \citet{tan2021progressive}.

\end{itemize}

We provide the same \textit{generated} content plans to the baselines to make the comparison fair. We use the generated \textit{theme} as input to the Fusion and Plan-and-write models, which is analogous to the prompt or the title. On the other hand, we extract keywords using TF-IDF following \citet{tan2021progressive} from all plot elements to prepare the input to ProGen. 

We also experiment with a baseline \textbf{GPT2-XL without content planning} where we sample a list of stories by providing the instruction ``\textbf{Task:} Write a plot summary.\textbackslash n \textbf{Plot summary:}''. We limit the story length to 150 tokens in all models for ease of human evaluation~\footnote{Additional experimental details can be found in Appendix~\ref{sec:appendix-details}.}.

\textbf{RQ1: Which story ending ranking performs better for ScratchPlot?}\quad We conduct a pair-wise comparison on the following story ending ranking methods: selecting the highest NSP (next sentence prediction) score, the lowest PPL (perplexity) score, and a random story and ending pair. For each pair-wise evaluation, we randomly sample 50 content plans where the two methods select different story endings~\footnote{They may or may not have the same story body.}. Each time, we present the annotators two stories in randomized order and ask the annotators to rate which story ends better. Finally, we take the majority vote from three annotators for each comparison and present the result in Table~\ref{tab:pairwise}. We also show randomly sampled stories and story endings selected by different methods in Table~\ref{tab:sample-ending}.

\begin{table}[!htbp]
\centering
  \begin{tabular}{lc}
    \hline
    \textbf{Method} & \textbf{Win:Lose} \\\hline
	NSP vs. \textbf{Random} & 19:31 \\
	\textbf{PPL} vs. Random\textbf{*} & 35:15 \\
	\textbf{PPL} vs. NSP \textbf{*} & 32:18 \\\hline
  \end{tabular}
  
\caption{Pair-wise comparison of the story ending ranking methods. The winning method in each comparison is highlighted in bold. * indicates statistical significance using two-sided Wilcoxon signed-rank test with $p$=0.05.}
\label{tab:pairwise}
\end{table}

\begin{table*}[h]
\centering
  \begin{tabularx}{\textwidth}{p{2.4cm}p{12.4cm}}
    \cline{1-2}
    \textbf{Location:} & San Francisco\\
    \textbf{Cast:} & John Jones \textit{and} Evelynn\\
    \textbf{Genre:} & dark fantasy\\
    \textbf{Theme:} & The specter of the future is in the telling \\
    \cline{1-2}
	\textbf{Random:} & The protagonist is a woman who has become a cyborg. She must discover who she really is while dealing with the problems of a modern metropolis. John Jones is a mysterious person who claims to be an assassin who is also an inventor and an expert on the cyborgs and other strange and mysterious beings. Evelynn Jones is an attractive and brilliant woman who finds herself caught between the two worlds, and the protagonist's own conflicted past. \texttt{[SEP]} He had to be the hero! It was a tragedy that he couldn't be. He was a killer, an assassin. He had to do it and he did it with a smile on his face.\\
	\cline{1-2}
	\textbf{NSP:} & As the year comes to a close, two people begin to lose their grip on reality. Evelynn's father has been dead for several years, but her mother never truly accepted her fate and continues to live a life of
denial. John Jones was once a successful businessman, but his fortune was spent by a mysterious corporation that he believed was his own. His life is about to take a turn for the worse as he discovers that Evelynn is living with a life-long secret. \texttt{[SEP]} In order for her to see the future, she'll have to take the risk. \\
	\cline{1-2}
	\textbf{PPL:} & In the past, John Jones used to be a normal person who worked for the government. But after a strange accident, he was taken to a secret facility, where he met the girl he loved, Evelynn, and started a relationship with her. But as the years passed, his memories started to grow more and more vague, and he started to realize that he didn't really remember how he got into that facility. \texttt{[SEP]} After a few months, John's memory returned to normal. He and Evelynn had their own children, but the memories remained.\\
	\cline{1-2}
  \end{tabularx}
  
\caption{Story body and ending selected by different algorithms. We manually insert a \texttt{[SEP]} token to indicate the boundary between the story body and ending.}
\label{tab:sample-ending}
\end{table*}

Based on the human rating, PPL selects more favorable story endings than NSP or Random. Surprisingly, NSP performs worse than random story and ending pairs. We hypothesize it might be due to the weakness of the NSP pre-training task. The negative examples during NSP pre-training are random sentences from the \textit{corpus}, which might be too trivial. Therefore, when the story ending and the story have word overlap, the model often predicts a very high $\mathcal{P}_{NSP}$ close to 1, causing the comparison to be unreliable.

Besides human evaluation on stories generated by PLMs, we also evaluate the story ending ranking methods on the Story Cloze Test dataset~\citep{mostafazadeh-etal-2016-corpus}. The dataset was created using crowd-sourcing to test models' commonsense story understanding. Each story contains four preceding sentences, a `right ending' and a `wrong ending'. The task is to predict which ending is the right one. \citet{mostafazadeh-etal-2016-corpus} instructed the crowdworkers to share at least one of the characters of the story in the ending and to ensure the ending sentence is entirely realistic and sensible when read in isolation. Therefore, the task is non-trivial and shallow techniques barely outperform a random baseline.

\begin{table}[!htbp]
\centering
  \begin{tabular}{lc}
    \hline
    \textbf{Method} & \textbf{Accuracy} \\\hline
    Word2Vec & 0.539 \\
    Skip-thought & 0.552 \\
    DSSM & 0.585 \\\hline
	NSP & 0.580 \\
	PPL & \textbf{0.587} \\\hline
  \end{tabular}
  
\caption{The accuracy of various models on the Story Cloze Test test dataset. We copied the results of the first three baselines from \citet{mostafazadeh-etal-2016-corpus}.}
\label{tab:cloze-test}
\end{table}

We report the Story Cloze Test accuracy in Table~\ref{tab:cloze-test}, along with the three best performing baselines reported in \citet{mostafazadeh-etal-2016-corpus}. Word2Vec~\citep{mikolov2013distributed} and Skip-thoughts~\citep{kiros2015skip} calculate the semantic representation using the average Word2Vec embedding and Sentence2Vec embedding separately. The models predict the story ending whose embedding is nearest to the preceding story's embedding. Deep Structured Semantic Model (DSSM)~\citep{huang2013learning} employs two jointly trained deep neural networks to project the preceding story context and the story ending into the same semantic space.

NSP and PPL performed comparably with DSSM. Notably, DSSM was trained on the full ROCStories dataset~\citep{mostafazadeh-etal-2016-corpus}, where the examples of the Story Cloze Test are drawn. It also used the Story Cloze Test validation set to perform hyper-parameter tuning. In contrast, NSP and PPL do not require any in-domain data or task-specific training. PPL's performance was better than NSP, consistent with the previous human evaluation results. Therefore, we use PPL to rank the story endings in subsequent experiments.

\textbf{RQ2: How does \textsc{S{\scriptsize CRATCH}P{\scriptsize LOT}} compare to the baselines?}\quad We generate 50 stories using each model and invite three crowdworkers to evaluate each story on the following fine-grained aspects: naturalness, interestingness, and cohesiveness. We take the average of the scores assigned by the annotators as the final score. Appendix~\ref{sec:appendix-human-eval} provides full details of the crowdsource evaluation. 

Table~\ref{tab:main-result} overviews the result, and Table~\ref{tab:sample-model} shows a randomly sampled content plan with stories generated by each model. We notice that the Fusion model tends to generate stories that consist primarily of dialogues, such as the example in Table~\ref{tab:sample-model}. It is also prone to generating repetitions.

\begin{table}[!htbp]
\centering
  \begin{tabular}{llll}
    \hline
    {\textbf{Model}} & \textbf{natur} & \textbf{inter} & \textbf{cohes} \\
    \hline
    Fusion & 2.13 & 2.31 & 1.89  \\
    Plan-and-write & 2.79 & 1.70 & 2.66 \\
    ProGen & 2.13 & 3.05 & 1.88  \\
    \hline
    \textsc{S{\scriptsize CRATCH}P{\scriptsize LOT}} & \textbf{4.04}* & \textbf{3.99}* & \textbf{3.47}  \\
    \; w/o content plan & 3.64 & 3.19 & 3.41  \\ \hline
  \end{tabular}
  
\caption{Human evaluation result of various models on different aspects. The columns denote naturalness, interestingness, cohesiveness. All scores are on a scale from 1 (worst) to 5 (best). Best scores for each aspect are highlighted in bold. * indicates statistical significance compared with the second best system using two-sided paired t-test with $p$=0.01.}
\label{tab:main-result}
\end{table}

Plan-and-write and ProGen both use a list of keywords as the content plan but are trained on different corpora. Plan-and-write generates short commonsense stories similar to the ROCStories dataset it is trained on. Thanks to its storyline generation, there is a logical sequence among the sentences. However, it lacks diversity in sentence structure. The stories also do not have rich plots and characters, causing them to have the lowest interestingness score among all baselines.

On the other hand, ProGen is trained on the {W{\scriptsize RITING}P{\scriptsize ROMPTS}} dataset, consisting of creative ``free-style'' stories. Unlike Plan-and-write, the content plan ProGen generates appears like a random bag of words. The stories are also often not logical. The comparison between the two models suggests that while a list of keywords is suitable as the content plan for short stories with mostly single-predicate sentences, it fails to generate cohesive stories with more nuance.

Compared to the baselines, \textsc{S{\scriptsize CRATCH}P{\scriptsize LOT}} performed best on all aspects, the improvement in interestingness being especially pronounced.

\textbf{RQ3: Does unsupervised content planning help?}\quad Different from previous work in story content planning, \textsc{S{\scriptsize CRATCH}P{\scriptsize LOT}} is entirely unsupervised, i.e., we prompt the same PLM that generates the story to generate the content plan without a need of finetuning.

We first measure the quality of the generated content plan by inviting an \textit{expert} annotator to rate all plot elements generated offline~\footnote{We describe the details of plot element generation in Appendix~\ref{sec:appendix-details}. Plot elements are much shorter and faster to rate. Therefore, we use an expert annotator for superior accuracy.}. We use binary rating (acceptable/unacceptable) for location, cast, and genre. We use a scale from 1 to 5 for theme because it is more subjective. Table~\ref{tab:plan-result} presents the average expert rating for each generated plot element. As we can see, the PLM generates simple fields like locations and person names with high quality. However, some generated themes are ambiguous or nonsensical.

\begin{table}[!htbp]
\centering
  \begin{tabular}{lcccc}
    \hline
    \textbf{Element} & Location & Cast & Genre & Theme \\
    \hline
    \textbf{Count} & 43 & 493 & 24 & 117 \\
    \textbf{Score} & 0.930 & 0.931 & 0.792 & 0.654 \\ \hline
  \end{tabular}
  
\caption{Expert rated scores for generated plot elements normalized to the range of 0 (worst) to 1 (perfect).}
\label{tab:plan-result}
\end{table}

\textsc{S{\scriptsize CRATCH}P{\scriptsize LOT}} outperformed the baseline without content planning in all aspects in Table~\ref{tab:main-result}, demonstrating the contribution of content plans in story generation even when they contain noise.

Furthermore, we measure \textit{intra}-story lexical diversity using self-BLEU~\citep{zhu2018texygen} and \textit{within}-story lexical diversity (or repetition) using \textit{distinct-n}~\citep{li-etal-2016-diversity} and summarize the result in Table~\ref{tab:diversity-result}. Unsurprisingly, the baseline without explicit content planning generates less diverse stories because they are sampled by conditioning on the same instruction. It also generates more \textit{within}-story repetitions than \textsc{S{\scriptsize CRATCH}P{\scriptsize LOT}}.

\begin{table}[!htbp]
\centering
  \begin{tabular}{lcccc}
    \hline
    \multirow{2}{*}{Model} & \multicolumn{2}{c}{self-BLEU} & \multicolumn{2}{c}{distinct-n} \\
    & $n$=1 & $n$=2 & $n$=1 & $n$=2 \\
    \hline
    Fusion & .805 & .483 & .079 & .316  \\
    Plan-and-write & .960 & .900 & .056 & .150 \\
    ProGen & .805 & \textbf{.347} & .191 & .683 \\
    \hline
    \textsc{S{\scriptsize CRATCH}P{\scriptsize LOT}} & \textbf{.782} & .353 & \textbf{.228} & \textbf{.705} \\
    \; w/o content plan & .799 & .380 & .204 & .650 \\ \hline
  \end{tabular}
  
\caption{Result of self-BLEU scores to measure intra-story diversity (the lower the better) and distinct-n scores to measure repetitions (the higher the better). The best results are highlighted in bold.}
\label{tab:diversity-result}
\end{table}

Qualitatively, Table~\ref{tab:sample-model} demonstrates that while the PLM without content planning started the story well, it began ``looping'' towards the end, a common problem referred to as neural text degeneration~\citep{holtzman2019curious}. In contrast, \textsc{S{\scriptsize CRATCH}P{\scriptsize LOT}} generates a story with natural progression.

Plan-and-write often ignores the input theme completely and predicts common storylines similar to its training data, such as ``wanted; decided; bought; started; happy'' and ``wanted; decided; months; started; felt''. Therefore, it has much worse inter-/intra-story diversity.

\begin{table*}[!htp]
\centering
  \begin{tabularx}{\textwidth}{p{2.8cm}p{12.4cm}}
    \cline{1-2}
    \textbf{Location:} & Alameda County\\
    \textbf{Cast:} & John \textit{and} Cheryl\\
    \textbf{Genre:} & literary fiction\\
    \textbf{Theme:} & The only problem with the house was the people. \\
    \cline{1-2}
	\textbf{Fusion:} & ``What?'' she whispered, as she pulled a small piece from her coat. It was a small piece of paper, with a picture of a young man sitting on that couch. `` What?'' he said, as he flipped the picture around. ``My name is John, and I 'm here to give you a story.'' he said, as the photo of his wife came out of his hand and he pulled out a book. The picture was of some kind of book, with the Briggs on it. ``What'' she said, as she looked at it . `` I don't know,'' John said, as he read the book. ``It was about to get pretty bad. I don't like it here.'' ``It's not good,'' she said. ``What did you tell me?'' he asked, as the picture was on the couch. ``I just can't believe it\\
	\cline{1-2}
	\textbf{Plan-and-write:} & \textbf{Content plan:} \textit{wanted; decided; bought; started; happy} \\
	& I wanted to go to the store. I decided to buy a new car. I bought a new car. I started to look for a new car. I was happy to have a new car.\\
	\cline{1-2}
	\textbf{ProGen:} &  \textbf{Content plan:} \textit{object; active; hemperforming; fail; blinked; first; son; joined; wife; bedroom; boys; excited; boy; jerry;} \\
	& The first son I held inside my wife skin and body, I rested my head on her chest and prayed to the waiting Forbiddone, loved by the boy and girl alike, and that the house would bring smiles to all. Jerry Fairey, anxious and excited, pushed his face into my naked flesh. Jerry sounded just like his daddy and chirped, ``Grr!'' and they laughed together. Jerry took a deep breath, thought of again, ``Fuck this. This place is empty, people.'' Jerry slept.\\
	\cline{1-2}
	\textbf{\textsc{S{\scriptsize CRATCH}P{\scriptsize LOT}}:} & John and Cheryl had been divorced for years, and while the divorce was final, there were still legal problems. John had bought a house in a nice part of town, and now, as it was approaching its five-year anniversary, his friends began asking questions about the house. The house was nice, and had nice rooms and a pool and all that. But the people who lived there had some problems. They lived with their mother and her family of five, with no money and no job. On the first day of the week of September, 1995, John and his friends got in their cars and drove back to the house they had been renting in the woods.\\
	\cline{1-2}
	\textbf{SP w/o content plan:} & An interesting tale of murder, betrayal, and the power of the human mind. (Summary by Ransom, written by L. Ron Hubbard.) Ransom: ``In the year 2392, an entire city has been destroyed and the entire population of a neighboring planet is now on the run. In this story, an ordinary man is caught up in a conspiracy that puts him on the edge of a world-changing discovery.'' (Plot summary by L Ron Hubbard, written to a friend.) Plot Summary: A plot summary of one of Hubbard's books. A summary from Hubbard's biography, Rise to Power. \\
	\cline{1-2}
  \end{tabularx}
  
\caption{Stories generated by different models. All models except for ``\textsc{S{\scriptsize CRATCH}P{\scriptsize LOT}} w/o content plan'' use the same plot element(s) listed above. We also show the intermediate content plans generated by Plan-and-write and ProGen.}
\label{tab:sample-model}
\end{table*}

\section{Related Work}
\label{sec:related-work}

\subsection{Story Content Planning}

\citet{fan-etal-2018-hierarchical} introduced hierarchical story generation by first generating a prompt then transforming it into a complete story. They proposed a novel fusion-based architecture to improve the relevance between the generation and the input prompt. They also collected a large dataset of 300K (prompt, story) pairs from Reddit’s {W{\scriptsize RITING}P{\scriptsize ROMPTS}} forum. The dataset has been widely used in subsequent work in story generation.

\citet{xu-etal-2018-skeleton} and \citet{yao2019plan} argued that Sequence-to-Sequence (seq2seq) models generate sentences autoregressively and are not good at modeling semantic dependencies \textit{among} sentences. Therefore, they proposed to use a list of keyphrases as the intermediate content plan. \citet{xu-etal-2018-skeleton} used policy gradient~\citep{sutton1999policy} to train the keyphrase extraction module, optimizing towards rewards from the generative module. \citet{yao2019plan} explored two strategies: dynamic schema that generates the next keyword in the content plan and the next sentence in the story at each step, and static schema that generates all keywords in the content plan and generates sentences conditioned on the \textit{complete} content plan. They empirically showed that the static schema performed better and conjectured that it generates more coherent stories because it plans the storyline holistically.

Similarly, \citet{tan2021progressive} used lists of keywords as the content plan. However, their proposed method, ProGen, is a multi-stage Transformers seq2seq model, extracting keywords at different granularities. Each stage takes the output from the previous stage and adds finer-grained details.

Moreover, other reprensentations have been used for story content planning. ~\citet{fan-etal-2019-strategies} and \citet{goldfarb-tarrant-etal-2020-content} utilized predicate-argument tuples extracted using Semantic Role Labeling. \citet{sun2020summarize} employed extractive summarization to generate paragraph summaries from stories as the content plan. \citet{shen-etal-2019-towards} used a hierarchically-structured Variational autoencoders~\citep{bowman-etal-2016-generating} to infer latent representations at word- and sentence-level. During inference, they generate a series of plan vectors before word-level realization. 

Unlike previous works, we use \textit{heterogenous} plot elements sampled from a PLM as the content plan (e.g., cast, location, genre). We also do not require any fine-tuning and rely solely on off-the-shelf PLMs.

\subsection{Story Ending Generation}

Previous work in story ending generation focused mostly on short commonsense stories. \citet{mostafazadeh-etal-2016-corpus} introduced ROCStories, a crowd-sourced corpus of 50k five-sentence commonsense stories. The corpus is limited to non-fictional daily life stories and focuses on being logically meaningful instead of dramatic and entertaining. \citet{mostafazadeh-etal-2016-corpus} also introduced the Story Cloze Test task, predicting the correct ending of sample stories from the ROCStories dataset. 

\citet{xu2020controllable} first extracted keywords from the story context in the ROCStories dataset, then retrieved relevant external knowledge from ConceptNet~\citep{speer-havasi-2012-representing}. Finally, they generated story endings conditioned on the story context and the retrieved knowledge.

\citet{ji2020language} argued that retrieving individual knowledge triples ignores the rich structure within the knowledge graph. To this end, they extracted sub-graphs using the story context and encoded them using a composition-based graph convolutional networks (GCN)~\citep{vashishth2019composition}. Finally, they performed multi-hop reasoning to generate the story ending.

\citet{rashkin-etal-2020-plotmachines} introduced a simpler approach to story ending generation. Their model, {P{\scriptsize LOT}M{\scriptsize ACHINES}}, added special discourse tokens to signal the introduction, body, and conclusion paragraphs in the story. The special token embeddings are trained with the model and help it to learn different writing styles of different parts of the story. 

Our story ending generation is most similar to \citet{rashkin-etal-2020-plotmachines} in that we do not perform explicit reasoning but rely on PLMs. However, different from \citet{rashkin-etal-2020-plotmachines}, we use natural language instructions instead of trainable embeddings to signal the model to end the story.

\citet{tan2021progressive} and \citet{sun2020summarize} used next sentence prediction (NSP) from BERT~\citep{devlin-etal-2019-bert} as an automatic metric to measure intra-sentence coherence. However, we demonstrated in RQ1 that the conditional perplexity score is a more reliable metric. Future work can consider using this metric to measure sentence-level coherence instead.

\section{Conclusion}
\label{sec:conclusion}

We introduced \textsc{S{\scriptsize CRATCH}P{\scriptsize LOT}}, a framework to perform unsupervised content planning for story generation using only pretrained language models (PLM). \textsc{S{\scriptsize CRATCH}P{\scriptsize LOT}} achieved strong results compared to supervised baselines fine-tuned on large parallel corpora and a PLM without access to content plans. In future work, we plan to generalize the framework to other types of long-form text.

\section*{Acknowledgements}

Yiping was supported by the scholarship from `The 100$^{th}$ Anniversary Chulalongkorn University Fund for Doctoral Scholarship' and also `The 90$^{th}$ Anniversary Chulalongkorn University Fund (Ratchadaphiseksomphot Endowment Fund)'. In addition, the crowdsource human evaluation was funded by Toloka Research Grant~\footnote{\url{https://toloka.ai/academy/grants}}. We appreciate their generosity and support for the research community. We would also like to thank the anonymous reviewers for their valuable feedback.


\section*{Ethical Considerations}
\label{sec:ethical}

Our proposed method is intended for creative text composition. The generated stories can be either consumed by readers or help writers to come up with new ideas. There are several potential risks if the proposed method is not deployed with care. However, they are inherent from large pre-trained language models (PLMs) instead of intrinsic to our method. 

First, PLMs may recall partially from the training data instead of composing stories from scratch. Due to the vast size of the pre-training data, it is not feasible to measure what percentage of the generated stories are ``original''. Secondly, the system sometimes generates real person names of famous people as the main characters. It should be noted that the system is for literature purposes and is not meant to be a factual report of real persons or anecdotes. Lastly, the system might generate inappropriate or disrespectful stories to a particular population, such as the genres ``biblical epic'' and ``erotica''. Manual curation or automatic content filtering can be deployed to mitigate this problem.

We relied on crowdworkers to conduct human evaluations in this work. The crowdworkers are from various countries, and the adequate payment differs drastically. Therefore, we target paying \$6.0 per hour. Some of the tasks took longer than we initially estimated, and we issued all crowdworkers a one-time bonus of \$0.2 to compensate.

Although we use a relatively large PLM (GPT2-XL; 1.5 billion parameters), our approach does not require training. Generating a single story takes around 1 minute, consuming 0.003 kWh power based on the max power consumption of the Quadro P5000 we used in the experiment.

\bibliography{anthology,custom}

\begin{thebibliography}{37}
\expandafter\ifx\csname natexlab\endcsname\relax\def\natexlab#1{#1}\fi

\bibitem[{Bird and Loper(2004)}]{bird-loper-2004-nltk}
Steven Bird and Edward Loper. 2004.
\newblock \href {https://aclanthology.org/P04-3031} {{NLTK}: The natural
  language toolkit}.
\newblock In \emph{Proceedings of the {ACL} Interactive Poster and
  Demonstration Sessions}, pages 214--217, Barcelona, Spain. Association for
  Computational Linguistics.

\bibitem[{Bosselut et~al.(2018)Bosselut, Celikyilmaz, He, Gao, Huang, and
  Choi}]{bosselut2018discourse}
Antoine Bosselut, Asli Celikyilmaz, Xiaodong He, Jianfeng Gao, Po-Sen Huang,
  and Yejin Choi. 2018.
\newblock \href {https://doi.org/10.18653/v1/N18-1016} {Discourse-aware neural
  rewards for coherent text generation}.
\newblock In \emph{Proceedings of the 2018 Conference of the North {A}merican
  Chapter of the Association for Computational Linguistics: Human Language
  Technologies, Volume 1 (Long Papers)}, pages 173--184, New Orleans,
  Louisiana. Association for Computational Linguistics.

\bibitem[{Bowman et~al.(2015)Bowman, Angeli, Potts, and
  Manning}]{bowman-etal-2015-large}
Samuel~R. Bowman, Gabor Angeli, Christopher Potts, and Christopher~D. Manning.
  2015.
\newblock \href {https://doi.org/10.18653/v1/D15-1075} {A large annotated
  corpus for learning natural language inference}.
\newblock In \emph{Proceedings of the 2015 Conference on Empirical Methods in
  Natural Language Processing}, pages 632--642, Lisbon, Portugal. Association
  for Computational Linguistics.

\bibitem[{Bowman et~al.(2016)Bowman, Vilnis, Vinyals, Dai, Jozefowicz, and
  Bengio}]{bowman-etal-2016-generating}
Samuel~R. Bowman, Luke Vilnis, Oriol Vinyals, Andrew Dai, Rafal Jozefowicz, and
  Samy Bengio. 2016.
\newblock \href {https://doi.org/10.18653/v1/K16-1002} {Generating sentences
  from a continuous space}.
\newblock In \emph{Proceedings of The 20th {SIGNLL} Conference on Computational
  Natural Language Learning}, pages 10--21, Berlin, Germany. Association for
  Computational Linguistics.

\bibitem[{Devlin et~al.(2019)Devlin, Chang, Lee, and
  Toutanova}]{devlin-etal-2019-bert}
Jacob Devlin, Ming-Wei Chang, Kenton Lee, and Kristina Toutanova. 2019.
\newblock \href {https://doi.org/10.18653/v1/N19-1423} {{BERT}: Pre-training of
  deep bidirectional transformers for language understanding}.
\newblock In \emph{Proceedings of the 2019 Conference of the North {A}merican
  Chapter of the Association for Computational Linguistics: Human Language
  Technologies, Volume 1 (Long and Short Papers)}, pages 4171--4186,
  Minneapolis, Minnesota. Association for Computational Linguistics.

\bibitem[{Fan et~al.(2018)Fan, Lewis, and Dauphin}]{fan-etal-2018-hierarchical}
Angela Fan, Mike Lewis, and Yann Dauphin. 2018.
\newblock \href {https://doi.org/10.18653/v1/P18-1082} {Hierarchical neural
  story generation}.
\newblock In \emph{Proceedings of the 56th Annual Meeting of the Association
  for Computational Linguistics (Volume 1: Long Papers)}, pages 889--898,
  Melbourne, Australia. Association for Computational Linguistics.

\bibitem[{Fan et~al.(2019)Fan, Lewis, and Dauphin}]{fan-etal-2019-strategies}
Angela Fan, Mike Lewis, and Yann Dauphin. 2019.
\newblock \href {https://doi.org/10.18653/v1/P19-1254} {Strategies for
  structuring story generation}.
\newblock In \emph{Proceedings of the 57th Annual Meeting of the Association
  for Computational Linguistics}, pages 2650--2660, Florence, Italy.
  Association for Computational Linguistics.

\bibitem[{Goldfarb-Tarrant et~al.(2020)Goldfarb-Tarrant, Chakrabarty,
  Weischedel, and Peng}]{goldfarb-tarrant-etal-2020-content}
Seraphina Goldfarb-Tarrant, Tuhin Chakrabarty, Ralph Weischedel, and Nanyun
  Peng. 2020.
\newblock \href {https://doi.org/10.18653/v1/2020.emnlp-main.351} {Content
  planning for neural story generation with aristotelian rescoring}.
\newblock In \emph{Proceedings of the 2020 Conference on Empirical Methods in
  Natural Language Processing (EMNLP)}, pages 4319--4338, Online. Association
  for Computational Linguistics.

\bibitem[{Holtzman et~al.(2019)Holtzman, Buys, Du, Forbes, and
  Choi}]{holtzman2019curious}
Ari Holtzman, Jan Buys, Li~Du, Maxwell Forbes, and Yejin Choi. 2019.
\newblock The curious case of neural text degeneration.
\newblock In \emph{Proceedings of the Seventh International Conference on
  Learning Representations}, New Orleans, USA.

\bibitem[{Hua and Wang(2020)}]{hua2020pair}
Xinyu Hua and Lu~Wang. 2020.
\newblock \href {https://doi.org/10.18653/v1/2020.emnlp-main.57} {{PAIR}:
  Planning and iterative refinement in pre-trained transformers for long text
  generation}.
\newblock In \emph{Proceedings of the 2020 Conference on Empirical Methods in
  Natural Language Processing (EMNLP)}, pages 781--793, Online. Association for
  Computational Linguistics.

\bibitem[{Huang et~al.(2013)Huang, He, Gao, Deng, Acero, and
  Heck}]{huang2013learning}
Po-Sen Huang, Xiaodong He, Jianfeng Gao, Li~Deng, Alex Acero, and Larry Heck.
  2013.
\newblock Learning deep structured semantic models for web search using
  clickthrough data.
\newblock In \emph{Proceedings of the 22nd ACM international conference on
  Information \& Knowledge Management}, pages 2333--2338, San Francisco, USA.

\bibitem[{Ji and Huang(2021)}]{ji2021discodvt}
Haozhe Ji and Minlie Huang. 2021.
\newblock \href {https://aclanthology.org/2021.emnlp-main.347} {{D}isco{DVT}:
  {G}enerating long text with discourse-aware discrete variational
  transformer}.
\newblock In \emph{Proceedings of the 2021 Conference on Empirical Methods in
  Natural Language Processing}, pages 4208--4224, Online and Punta Cana,
  Dominican Republic. Association for Computational Linguistics.

\bibitem[{Ji et~al.(2020)Ji, Ke, Huang, Wei, Zhu, and Huang}]{ji2020language}
Haozhe Ji, Pei Ke, Shaohan Huang, Furu Wei, Xiaoyan Zhu, and Minlie Huang.
  2020.
\newblock \href {https://doi.org/10.18653/v1/2020.emnlp-main.54} {Language
  generation with multi-hop reasoning on commonsense knowledge graph}.
\newblock In \emph{Proceedings of the 2020 Conference on Empirical Methods in
  Natural Language Processing (EMNLP)}, pages 725--736, Online. Association for
  Computational Linguistics.

\bibitem[{Kiros et~al.(2015)Kiros, Zhu, Salakhutdinov, Zemel, Urtasun,
  Torralba, and Fidler}]{kiros2015skip}
Ryan Kiros, Yukun Zhu, Russ~R Salakhutdinov, Richard Zemel, Raquel Urtasun,
  Antonio Torralba, and Sanja Fidler. 2015.
\newblock Skip-thought vectors.
\newblock \emph{Advances in neural information processing systems}, 28.

\bibitem[{Lee et~al.(2021)Lee, Bang, Madotto, and Fung}]{lee-etal-2021-towards}
Nayeon Lee, Yejin Bang, Andrea Madotto, and Pascale Fung. 2021.
\newblock \href {https://doi.org/10.18653/v1/2021.naacl-main.158} {Towards
  few-shot fact-checking via perplexity}.
\newblock In \emph{Proceedings of the 2021 Conference of the North American
  Chapter of the Association for Computational Linguistics: Human Language
  Technologies}, pages 1971--1981, Online. Association for Computational
  Linguistics.

\bibitem[{Lee et~al.(2020)Lee, Li, Wang, Yih, Ma, and
  Khabsa}]{lee-etal-2020-language}
Nayeon Lee, Belinda~Z. Li, Sinong Wang, Wen-tau Yih, Hao Ma, and Madian Khabsa.
  2020.
\newblock \href {https://doi.org/10.18653/v1/2020.fever-1.5} {Language models
  as fact checkers?}
\newblock In \emph{Proceedings of the Third Workshop on Fact Extraction and
  VERification (FEVER)}, pages 36--41, Online. Association for Computational
  Linguistics.

\bibitem[{Lewis et~al.(2020)Lewis, Liu, Goyal, Ghazvininejad, Mohamed, Levy,
  Stoyanov, and Zettlemoyer}]{lewis-etal-2020-bart}
Mike Lewis, Yinhan Liu, Naman Goyal, Marjan Ghazvininejad, Abdelrahman Mohamed,
  Omer Levy, Veselin Stoyanov, and Luke Zettlemoyer. 2020.
\newblock \href {https://doi.org/10.18653/v1/2020.acl-main.703} {{BART}:
  Denoising sequence-to-sequence pre-training for natural language generation,
  translation, and comprehension}.
\newblock In \emph{Proceedings of the 58th Annual Meeting of the Association
  for Computational Linguistics}, pages 7871--7880, Online. Association for
  Computational Linguistics.

\bibitem[{Li et~al.(2016)Li, Galley, Brockett, Gao, and
  Dolan}]{li-etal-2016-diversity}
Jiwei Li, Michel Galley, Chris Brockett, Jianfeng Gao, and Bill Dolan. 2016.
\newblock \href {https://doi.org/10.18653/v1/N16-1014} {A diversity-promoting
  objective function for neural conversation models}.
\newblock In \emph{Proceedings of the 2016 Conference of the North {A}merican
  Chapter of the Association for Computational Linguistics: Human Language
  Technologies}, pages 110--119, San Diego, California. Association for
  Computational Linguistics.

\bibitem[{Mikolov et~al.(2013)Mikolov, Sutskever, Chen, Corrado, and
  Dean}]{mikolov2013distributed}
Tomas Mikolov, Ilya Sutskever, Kai Chen, Greg~S Corrado, and Jeff Dean. 2013.
\newblock Distributed representations of words and phrases and their
  compositionality.
\newblock \emph{Advances in neural information processing systems}, 26.

\bibitem[{Mostafazadeh et~al.(2016)Mostafazadeh, Chambers, He, Parikh, Batra,
  Vanderwende, Kohli, and Allen}]{mostafazadeh-etal-2016-corpus}
Nasrin Mostafazadeh, Nathanael Chambers, Xiaodong He, Devi Parikh, Dhruv Batra,
  Lucy Vanderwende, Pushmeet Kohli, and James Allen. 2016.
\newblock \href {https://doi.org/10.18653/v1/N16-1098} {A corpus and cloze
  evaluation for deeper understanding of commonsense stories}.
\newblock In \emph{Proceedings of the 2016 Conference of the North {A}merican
  Chapter of the Association for Computational Linguistics: Human Language
  Technologies}, pages 839--849, San Diego, California. Association for
  Computational Linguistics.

\bibitem[{Radford et~al.(2019)Radford, Wu, Child, Luan, Amodei, Sutskever
  et~al.}]{radford2019language}
Alec Radford, Jeffrey Wu, Rewon Child, David Luan, Dario Amodei, Ilya
  Sutskever, et~al. 2019.
\newblock Language models are unsupervised multitask learners.
\newblock \emph{OpenAI blog}, 1(8):9.

\bibitem[{Rashkin et~al.(2020)Rashkin, Celikyilmaz, Choi, and
  Gao}]{rashkin-etal-2020-plotmachines}
Hannah Rashkin, Asli Celikyilmaz, Yejin Choi, and Jianfeng Gao. 2020.
\newblock \href {https://doi.org/10.18653/v1/2020.emnlp-main.349}
  {{P}lot{M}achines: Outline-conditioned generation with dynamic plot state
  tracking}.
\newblock In \emph{Proceedings of the 2020 Conference on Empirical Methods in
  Natural Language Processing (EMNLP)}, pages 4274--4295, Online. Association
  for Computational Linguistics.

\bibitem[{Reiter and Dale(1997)}]{reiter1997building}
Ehud Reiter and Robert Dale. 1997.
\newblock Building applied natural language generation systems.
\newblock \emph{Natural Language Engineering}, 3(1):57--87.

\bibitem[{Schick and Sch{\"u}tze(2021)}]{schick-schutze-2021-generating}
Timo Schick and Hinrich Sch{\"u}tze. 2021.
\newblock \href {https://aclanthology.org/2021.emnlp-main.555} {Generating
  datasets with pretrained language models}.
\newblock In \emph{Proceedings of the 2021 Conference on Empirical Methods in
  Natural Language Processing}, pages 6943--6951, Online and Punta Cana,
  Dominican Republic. Association for Computational Linguistics.

\bibitem[{Schick et~al.(2021)Schick, Udupa, and Sch{\"u}tze}]{schick2021self}
Timo Schick, Sahana Udupa, and Hinrich Sch{\"u}tze. 2021.
\newblock Self-diagnosis and self-debiasing: A proposal for reducing
  corpus-based bias in nlp.
\newblock \emph{arXiv preprint arXiv:2103.00453}.

\bibitem[{See et~al.(2019)See, Pappu, Saxena, Yerukola, and
  Manning}]{see2019massively}
Abigail See, Aneesh Pappu, Rohun Saxena, Akhila Yerukola, and Christopher~D.
  Manning. 2019.
\newblock \href {https://doi.org/10.18653/v1/K19-1079} {Do massively pretrained
  language models make better storytellers?}
\newblock In \emph{Proceedings of the 23rd Conference on Computational Natural
  Language Learning (CoNLL)}, pages 843--861, Hong Kong, China. Association for
  Computational Linguistics.

\bibitem[{Shen et~al.(2019)Shen, Celikyilmaz, Zhang, Chen, Wang, Gao, and
  Carin}]{shen-etal-2019-towards}
Dinghan Shen, Asli Celikyilmaz, Yizhe Zhang, Liqun Chen, Xin Wang, Jianfeng
  Gao, and Lawrence Carin. 2019.
\newblock \href {https://doi.org/10.18653/v1/P19-1200} {Towards generating long
  and coherent text with multi-level latent variable models}.
\newblock In \emph{Proceedings of the 57th Annual Meeting of the Association
  for Computational Linguistics}, pages 2079--2089, Florence, Italy.
  Association for Computational Linguistics.

\bibitem[{Speer and Havasi(2012)}]{speer-havasi-2012-representing}
Robyn Speer and Catherine Havasi. 2012.
\newblock \href
  {http://www.lrec-conf.org/proceedings/lrec2012/pdf/1072_Paper.pdf}
  {Representing general relational knowledge in {C}oncept{N}et 5}.
\newblock In \emph{Proceedings of the Eighth International Conference on
  Language Resources and Evaluation ({LREC}'12)}, pages 3679--3686, Istanbul,
  Turkey. European Language Resources Association (ELRA).

\bibitem[{Sun et~al.(2020)Sun, Fan, Sun, Meng, Wu, and Li}]{sun2020summarize}
Xiaofei Sun, Chun Fan, Zijun Sun, Yuxian Meng, Fei Wu, and Jiwei Li. 2020.
\newblock Summarize, outline, and elaborate: Long-text generation via
  hierarchical supervision from extractive summaries.
\newblock \emph{arXiv preprint arXiv:2010.07074}.

\bibitem[{Sutton et~al.(1999)Sutton, McAllester, Singh, and
  Mansour}]{sutton1999policy}
Richard~S Sutton, David McAllester, Satinder Singh, and Yishay Mansour. 1999.
\newblock Policy gradient methods for reinforcement learning with function
  approximation.
\newblock \emph{Advances in neural information processing systems}, 12.

\bibitem[{Tan et~al.(2021)Tan, Yang, Al-Shedivat, Xing, and
  Hu}]{tan2021progressive}
Bowen Tan, Zichao Yang, Maruan Al-Shedivat, Eric Xing, and Zhiting Hu. 2021.
\newblock \href {https://doi.org/10.18653/v1/2021.naacl-main.341} {Progressive
  generation of long text with pretrained language models}.
\newblock In \emph{Proceedings of the 2021 Conference of the North American
  Chapter of the Association for Computational Linguistics: Human Language
  Technologies}, pages 4313--4324, Online. Association for Computational
  Linguistics.

\bibitem[{Vashishth et~al.(2019)Vashishth, Sanyal, Nitin, and
  Talukdar}]{vashishth2019composition}
Shikhar Vashishth, Soumya Sanyal, Vikram Nitin, and Partha Talukdar. 2019.
\newblock Composition-based multi-relational graph convolutional networks.
\newblock In \emph{International Conference on Learning Representations}, New
  Orleans, USA.

\bibitem[{Wolf et~al.(2020)Wolf, Debut, Sanh, Chaumond, Delangue, Moi, Cistac,
  Rault, Louf, Funtowicz, Davison, Shleifer, von Platen, Ma, Jernite, Plu, Xu,
  Scao, Gugger, Drame, Lhoest, and Rush}]{wolf-etal-2020-transformers}
Thomas Wolf, Lysandre Debut, Victor Sanh, Julien Chaumond, Clement Delangue,
  Anthony Moi, Pierric Cistac, Tim Rault, Rémi Louf, Morgan Funtowicz, Joe
  Davison, Sam Shleifer, Patrick von Platen, Clara Ma, Yacine Jernite, Julien
  Plu, Canwen Xu, Teven~Le Scao, Sylvain Gugger, Mariama Drame, Quentin Lhoest,
  and Alexander~M. Rush. 2020.
\newblock \href {https://www.aclweb.org/anthology/2020.emnlp-demos.6}
  {Transformers: State-of-the-art natural language processing}.
\newblock In \emph{Proceedings of the 2020 Conference on Empirical Methods in
  Natural Language Processing: System Demonstrations}, pages 38--45, Online.
  Association for Computational Linguistics.

\bibitem[{Xu et~al.(2018)Xu, Ren, Zhang, Zeng, Cai, and
  Sun}]{xu-etal-2018-skeleton}
Jingjing Xu, Xuancheng Ren, Yi~Zhang, Qi~Zeng, Xiaoyan Cai, and Xu~Sun. 2018.
\newblock \href {https://doi.org/10.18653/v1/D18-1462} {A skeleton-based model
  for promoting coherence among sentences in narrative story generation}.
\newblock In \emph{Proceedings of the 2018 Conference on Empirical Methods in
  Natural Language Processing}, pages 4306--4315, Brussels, Belgium.
  Association for Computational Linguistics.

\bibitem[{Xu et~al.(2020)Xu, Patwary, Shoeybi, Puri, Fung, Anandkumar, and
  Catanzaro}]{xu2020controllable}
Peng Xu, Mostofa Patwary, Mohammad Shoeybi, Raul Puri, Pascale Fung, Anima
  Anandkumar, and Bryan Catanzaro. 2020.
\newblock \href {https://doi.org/10.18653/v1/2020.emnlp-main.226}
  {{MEGATRON}-{CNTRL}: Controllable story generation with external knowledge
  using large-scale language models}.
\newblock In \emph{Proceedings of the 2020 Conference on Empirical Methods in
  Natural Language Processing (EMNLP)}, pages 2831--2845, Online. Association
  for Computational Linguistics.

\bibitem[{Yao et~al.(2019)Yao, Peng, Weischedel, Knight, Zhao, and
  Yan}]{yao2019plan}
Lili Yao, Nanyun Peng, Ralph Weischedel, Kevin Knight, Dongyan Zhao, and Rui
  Yan. 2019.
\newblock Plan-and-write: Towards better automatic storytelling.
\newblock In \emph{Proceedings of the AAAI Conference on Artificial
  Intelligence}, volume~33, pages 7378--7385, Honolulu, USA.

\bibitem[{Zhu et~al.(2018)Zhu, Lu, Zheng, Guo, Zhang, Wang, and
  Yu}]{zhu2018texygen}
Yaoming Zhu, Sidi Lu, Lei Zheng, Jiaxian Guo, Weinan Zhang, Jun Wang, and Yong
  Yu. 2018.
\newblock Texygen: A benchmarking platform for text generation models.
\newblock In \emph{The 41st International ACM SIGIR Conference on Research \&
  Development in Information Retrieval}, pages 1097--1100.

\end{thebibliography}
\bibliographystyle{acl_natbib}


\appendix

\section{Full List of Task Descriptions}
\label{sec:appendix-td}

Table~\ref{tab:list-instructions} shows the complete list of task descriptions to generate various plot elements for content planning. 

\begin{table*}[!htp]
\centering
  \begin{tabularx}{\textwidth}{p{1.2cm}p{13.6cm}}
    \cline{1-2}
    \textbf{Element} & \textbf{Task Description}\\
    \cline{1-2}
    Location & \textbf{Task:} Write the name of a country.\textbackslash n \textbf{Country:} ``\newline
       		   \textbf{Task:} Write the name of a province.\textbackslash n \textbf{Province:} ``\newline
       		   \textbf{Task:} Write the name of a city.\textbackslash n \textbf{City:} ``\newline
       		   \textbf{Task:} Write the name of a county.\textbackslash n \textbf{County:} ``
       		   \\\cline{1-2}
	Cast & 	   \textbf{Task:} Write the male character's full name in a story that happened in \textsc{<x{\scriptsize 1}>}.
\textbf{Full name:} ``\newline
       		   \textbf{Task:} Write the female character's full name in a story that happened in \textsc{<x{\scriptsize 1}>}.
\textbf{Full name:} ``
       		   \\\cline{1-2}
    Genre &    \textbf{Task:} Write a story genre.\textbackslash n \textbf{Story genre:} ``\newline
    		   \textbf{Task:} Write a literary genre.\textbackslash n \textbf{Literary genre:} ``\newline
    		   \textbf{Task:} Write a novel genre.\textbackslash n \textbf{Novel genre:} ``
       		   \\\cline{1-2}
	Theme &    \textbf{Task:} Write the main point from a \textsc{<x{\scriptsize 1}>} story.\textbackslash n \textbf{Main point:} ``\newline
			   \textbf{Task:} Write the twist in a \textsc{<x{\scriptsize 1}>} story.\textbackslash n \textbf{Twist:} ``\newline
			   \textbf{Task:} Write the lesson learned from a \textsc{<x{\scriptsize 1}>} story.\textbackslash n \textbf{Lesson learned:} ``\newline
			   \textbf{Task:} Write the spectacle of a \textsc{<x{\scriptsize 1}>} story.\textbackslash n \textbf{Spectacle:} ``
       		   \\\cline{1-2}
  \end{tabularx}
  
\caption{Full list of task descriptions to generate each element. \textsc{<x{\scriptsize 1}>} denotes the previously generated element. Story body and story ending generation both use a single task description as shown in Figure~\ref{fig:overview}.}
\label{tab:list-instructions}
\end{table*}

\section{Post-Processing}
\label{sec:appendix-postprocessing}

We perform various post-processing depending on the plot elements. We rely on simple heuristics based on common errors we observe.

\textbf{Including tailing punctuations}\quad For some plot elements, we expect a phrase instead of a whole sentence. However, the PLM sometimes inserts a punctuation mark, such as a full stop or a comma. Therefore, we recursively remove punctuations at the end till the last character is a letter.

\textbf{Repeating the prompt}\quad We observe that the PLM sometimes repeats or rephrases the task description instead of trying to perform the task. Therefore, we filter out continuations that contain \textit{any} word in the task description (excluding stop words and the text replacing the placeholder \textsc{<x{\scriptsize 1}>}).

\textbf{Generating 1st or 2nd person pronouns}\quad We usually do not expect first or second-person pronouns in a story plot. When the model generates plot elements containing first or second-person pronouns, it is often generic or opinionated, such as ``I'll try not to think about it'' or ``You will not fail me.'' Therefore, we filter continuations containing a first or second-person pronoun of any case.

\textbf{Ignoring task description}\quad When generating the story, we want to ensure that it includes essential plot elements specified in the task description. Therefore, we filter out a story if it contains fewer than two of the following \{male character's first name, female character's first name, location\}.

Table~\ref{tab:list-postprocessings} overviews the post-processing applied when generating each type of output.

\begin{table*}[!htp]
\centering
  \begin{tabularx}{\textwidth}{lcccccc}
    \cline{1-7}
    \textbf{Post-processing} & \textbf{Location} & \textbf{Cast} & \textbf{Genre} & \textbf{Theme} & \textbf{Body} & \textbf{Ending}\\
    \cline{1-7}
    Remove tailing punctuations & \CheckmarkBold & \CheckmarkBold & \CheckmarkBold &  &  &  \\\cline{1-7}
	Filter repeating prompt & \CheckmarkBold & \CheckmarkBold & \CheckmarkBold & \CheckmarkBold & \CheckmarkBold & \CheckmarkBold \\\cline{1-7}  
	Filter 1st \& 2nd person pronouns &  & &  & \CheckmarkBold & \CheckmarkBold & \CheckmarkBold \\\cline{1-7}  
	Filter by plot elements &  &  &  &  & \CheckmarkBold &  \\\cline{1-7}  
  \end{tabularx}
  
\caption{Post-processing steps applied for each generation task.}
\label{tab:list-postprocessings}
\end{table*}

\section{Additional Experimental Setups}
\label{sec:appendix-details}

During training/generation, we use the tokenizers associated with the corresponding PLM in the HuggingFace library~\citep{wolf-etal-2020-transformers}. When calculating diversity and repetition, we use NLTK~\citep{bird-loper-2004-nltk} to perform word tokenization. We calculate self-BLEU scores using NLTK's \texttt{sentence\_bleu} method by treating each example as the reference in each round and averaging the BLEU scores over the whole dataset.

All experiments in this work are conducted on cloud instances with an NVIDIA Quadro P5000 GPU (16GB vRAM). The time to generate a story is roughly 1 minute, which includes generating multiple story bodies and endings and using scoring models to select the best candidate. Since we do not require any fine-tuning, using a CPU to perform inference is also possible. The reader can consider using a smaller GPT2-medium PLM instead of GPT2-XL when the resource is limited. The generation quality is comparable based on our observation.

\section{Details of Crowdsource Evaluation}
\label{sec:appendix-human-eval}

We conducted the crowdsource evaluations on the Toloka platform~\footnote{\url{https://toloka.ai/}}. In this section, we detail the specification of the annotation tasks, the quality control measures, and the stats of the annotation.

\subsection{Annotation Task Specifications}

For the fine-grained evaluation, we decompose it into a separate annotation task per aspect so that the annotators can focus on evaluating a single aspect and avoid context switching.

\textbf{Fine-grained evaluation}\quad Rate each story in the following aspects on a scale of 1 (worst) to 5 (best).

\begin{itemize}
  \item \textbf{Naturalness:} Is the story fluent and understandable? The language should be natural. Minor grammatical errors are acceptable if they do not affect understanding the story.
  \item \textbf{Interestingness:} Is the story interesting to readers? Rate this aspect as objective as possible. Assuming someone familiar with the particular genre, will the story interest them?
  \item \textbf{Cohesiveness:} Is the story cohesive and logical? Common problems include mixing up the characters and introducing illogical event sequences (unless it appears like a deliberate choice). 
\end{itemize} 

\begin{figure*}[!htbp]
\centering \includegraphics[width=418pt]{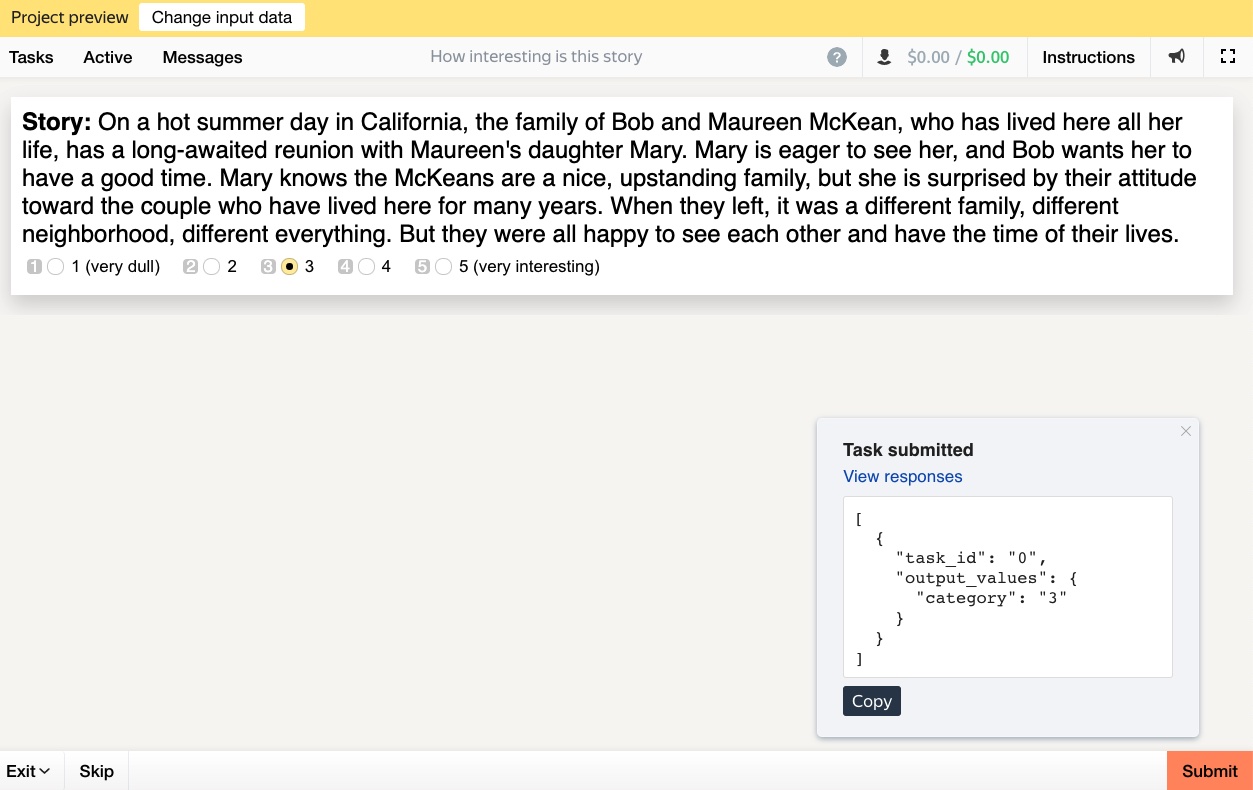}
  \caption{Annotation interface for the interestingness aspect. The interface for other aspects are analogous and we omit them for brevity.}  
  \label{fig:annotation-interface}
\end{figure*}

\begin{figure*}[!htbp]
\centering \includegraphics[width=450pt]{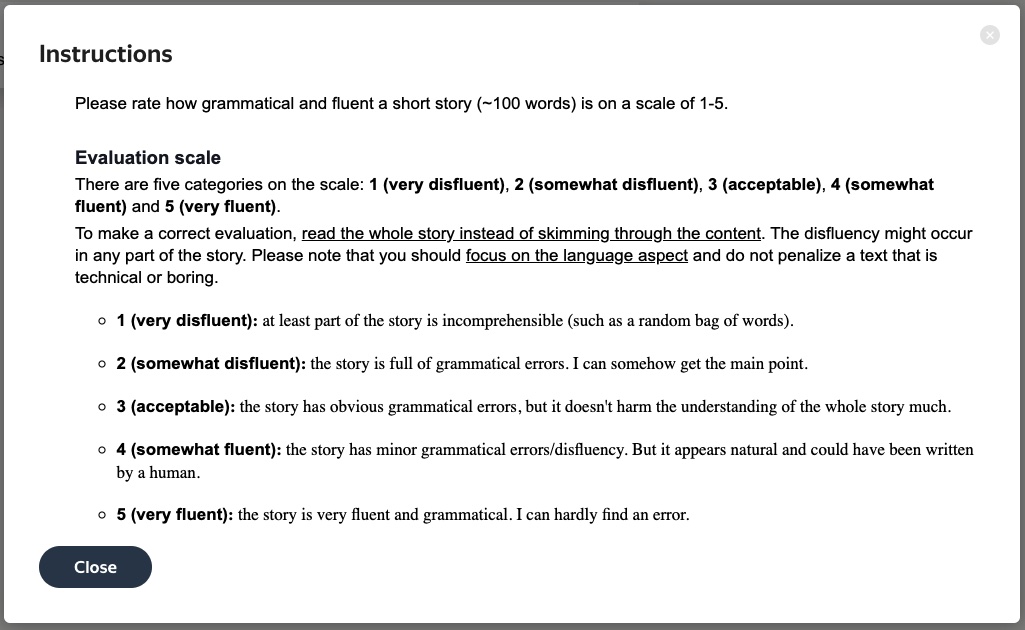}
  \caption{Annotation instructions for the naturalness aspect.}  
  \label{fig:annotation-instructions-naturalness}
\end{figure*}

\begin{figure*}[!htbp]
\centering \includegraphics[width=450pt]{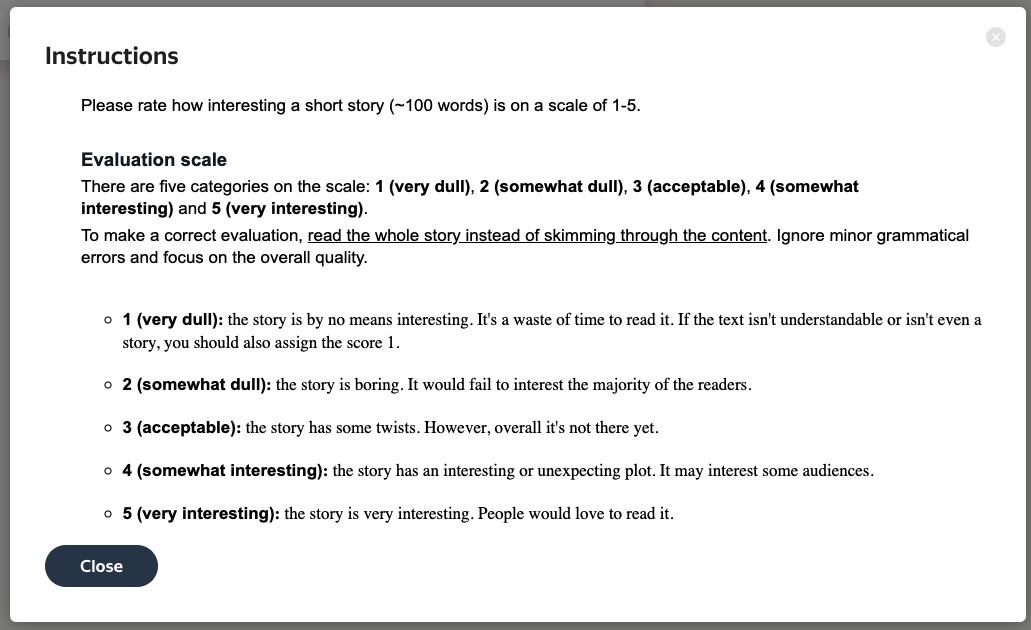}
  \caption{Annotation instructions for the interestingness aspect.}  
  \label{fig:annotation-instructions-interestingness}
\end{figure*}

\begin{figure*}[!htbp]
\centering \includegraphics[width=450pt]{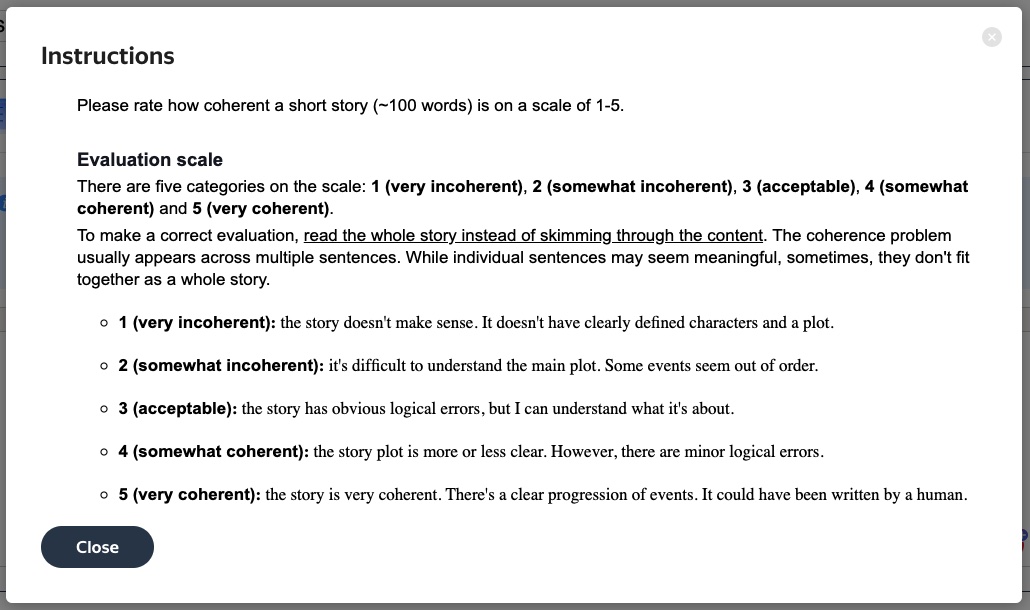}
  \caption{Annotation instructions for the cohesiveness aspect.}  
  \label{fig:annotation-instructions-coherence}
\end{figure*}

Figure~\ref{fig:annotation-interface} shows the annotation interface and Figure~\ref{fig:annotation-instructions-naturalness}, \ref{fig:annotation-instructions-interestingness}, \ref{fig:annotation-instructions-coherence} shows the detailed annotation instructions.

\begin{figure}[!htbp]
\centering \includegraphics[width=218pt]{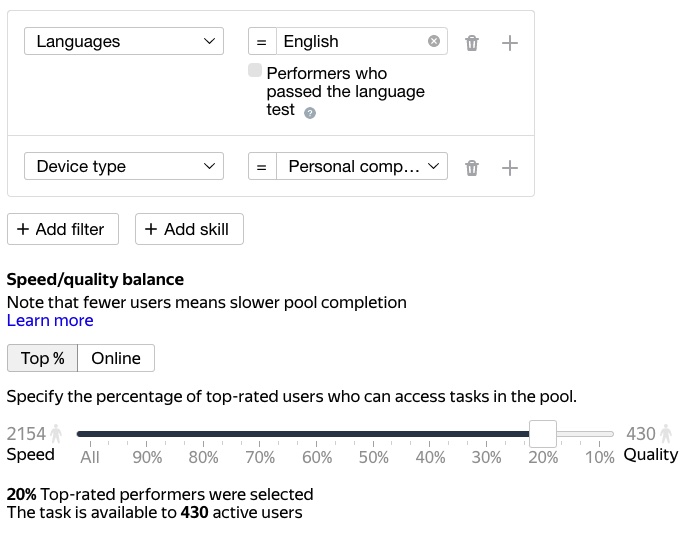}
  \caption{Audience filter for the annotation task pool.}  
  \label{fig:annotation-filter}
\end{figure}

\begin{figure*}[!htbp]
\centering \includegraphics[width=450pt]{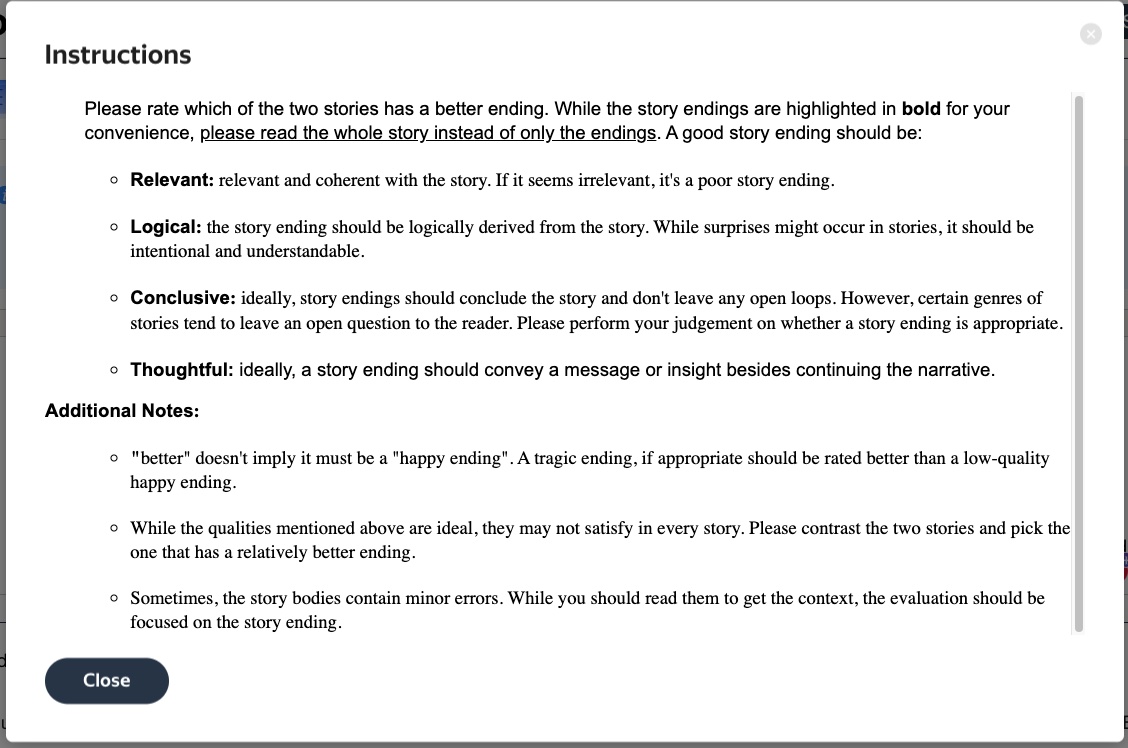}
  \caption{Annotation instructions for the story ending evaluation.}  
  \label{fig:annotation-instructions-pairwise}
\end{figure*}

\begin{figure*}[!htbp]
\centering \includegraphics[width=450pt]{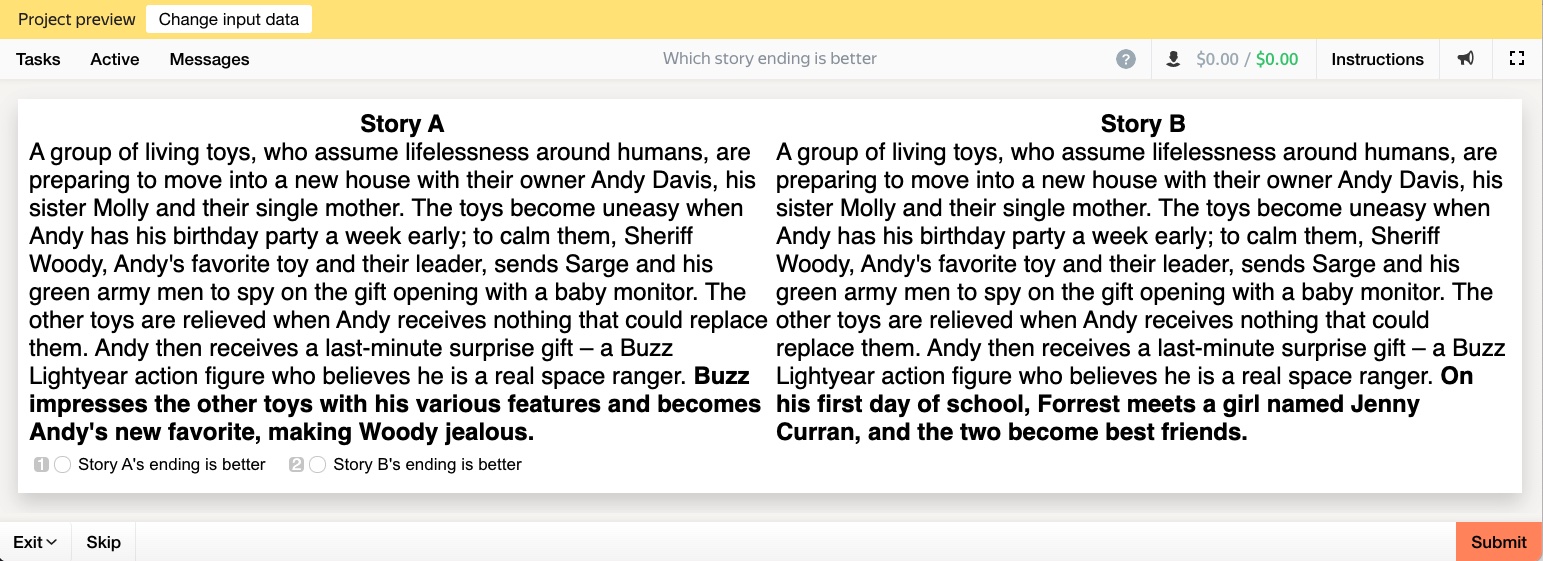}
  \caption{Annotation interface for the pair-wise story ending evaluation.}  
  \label{fig:annotation-interface-pairwise}
\end{figure*}

\textbf{Story ending evaluation}\quad Indicate which of the two stories has a better ending. A good story ending should be relevant to the story, logical, conclusive, and thoughtful. Figure~\ref{fig:annotation-instructions-pairwise} shows the full annotation instructions and Figure~\ref{fig:annotation-interface-pairwise} shows the annotation UI.

\subsection{Quality Control}

We select crowdworkers who are fluent in English and among the 20\% top-rated performers. Figure~\ref{fig:annotation-filter} shows a screenshot of the annotator filter. Additionally, they have to pass a short training session and correctly answer 3 out of 4 training questions to be selected for the main evaluation. 


During annotation, we apply various quality control rules, including limiting each annotator to no more than 50 tasks, adding occasional captcha to block bots, banning users who consistently submit tasks too fast  (less than 5 seconds for fine-grained evaluation and less than 10 seconds for story ending evaluation), and banning users who skip more than 5 tasks in a row.

\subsection{Annotation Task Stats}

\begin{figure*}[!htbp]
\centering \includegraphics[width=450pt]{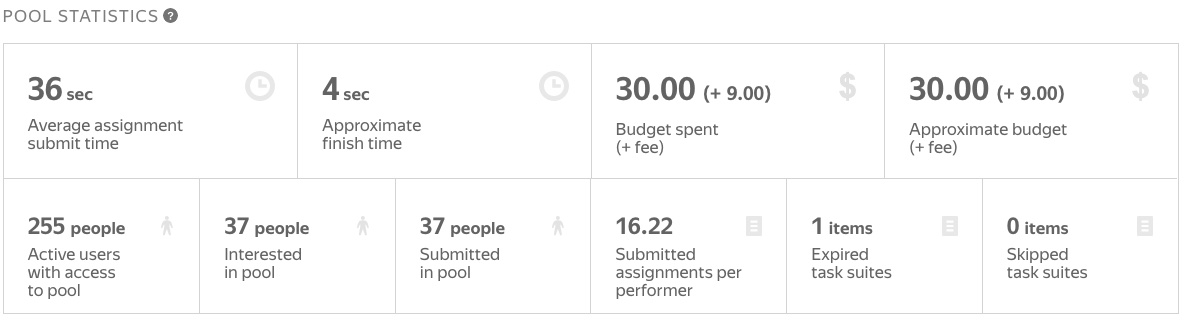}
  \caption{The annotation pool stats for the naturalness evaluation.}  
  \label{fig:pool-stats-naturalness}
\end{figure*}

We paid \$0.05 for each fine-grained evaluation task. On average, it took around 30 seconds to complete each task, making the average earning \$6 an hour. There are around 40 crowdworkers evaluating for each aspect. Figure~\ref{fig:pool-stats-naturalness} shows an example pool stats for the naturalness evaluation.

We paid \$0.1 for each story ending evaluation task, which takes on average 1 minute 13 seconds to complete. There are in total 20 crowdworkers participating in this evaluation task. 

The overall budget we spent on all crowdsource evaluations is \$300 (including payment and bonus to crowdworkers and platform fees).

\end{document}